\documentclass[11pt]{article}

\usepackage[a4paper,margin=2.54cm]{geometry}

\usepackage{amsmath,amssymb,amsfonts}

\usepackage{graphicx}
\usepackage{array}
\usepackage{siunitx}
\usepackage{adjustbox}
\usepackage{multirow}
\usepackage{booktabs}
\usepackage[table]{xcolor}
\usepackage{tabularx}
\usepackage{placeins}
\usepackage{algorithm}
\usepackage{algorithmic}
\usepackage{caption}
\usepackage{cite}

\usepackage{textcomp}
\usepackage[hidelinks]{hyperref}

\usepackage{microtype}

\usepackage{authblk}

\title{\bfseries ARCOS: Zero-shot Boundary Localization for Corneal Layer Segmentation Across Optical Coherence Tomography Devices}

\author[1]{Nuno Vivas Brás}
\author[2]{Benjamin Memmi}
\author[1]{Maëlle Bouhassane}
\author[2]{Cristina Georgeon}
\author[2]{Vincent Borderie}
\author[1]{Karsten Plamann}
\author[1]{Anatole Chessel}

\affil[1]{
Laboratoire d'Optique et Biosciences, CNRS, INSERM,
École polytechnique, Institut Polytechnique de Paris,
Palaiseau, France
}

\affil[2]{
GRC 32, Transplantation et Thérapies Innovantes de la Cornée,
Sorbonne Université,
Centre Hospitalier National d'Ophtalmologie des Quinze-Vingts,
Paris, France
}

\date{}

\begin{document}

\maketitle

\begin{abstract}
Accurate segmentation of corneal layers in optical coherence tomography (OCT) is essential for quantitative assessment of corneal morphology, including layer thickness and structural changes associated with disease or surgery. However, automatic segmentation remains challenging because corneal interfaces are thin, affected by speckle noise, and variable across acquisition devices. In this work, we propose ARCOS, a patch-based zero-shot boundary localization framework for corneal layer segmentation in clinical anterior-segment OCT images. Rather than performing conventional region classification, the method predicts boundary heatmaps for the main corneal interfaces from overlapping native-resolution patches. Patch-level predictions are stitched across the full B-scan and converted into boundary locations to obtain continuous, anatomically ordered layer segmentations. The network combines multi-scale feature fusion with a self-conditioned refinement module that uses intermediate boundary information to improve local heatmap predictions while preserving spatial detail. The method was evaluated on clinical OCT images acquired from multiple devices and compared with representative segmentation baselines using boundary localization and derived thickness metrics. The proposed method achieved an off-by-one boundary localization accuracy of 95.1\% and a mean absolute boundary error of 0.514 pixels on the matched-device test set.
In zero-shot cross-device evaluation, it maintained an average off-by-one accuracy of 84.3\% and a mean absolute boundary error of 0.855 pixels across unseen acquisition devices, outperforming the baseline models.
Thickness estimates derived from the predicted boundaries showed low error across corneal regions, supporting the method's use for quantitative corneal OCT analysis.
\end{abstract}

\section{Introduction}
\label{sec:introduction}

\begin{figure*}[!h]
\centerline{\includegraphics[width=\textwidth]{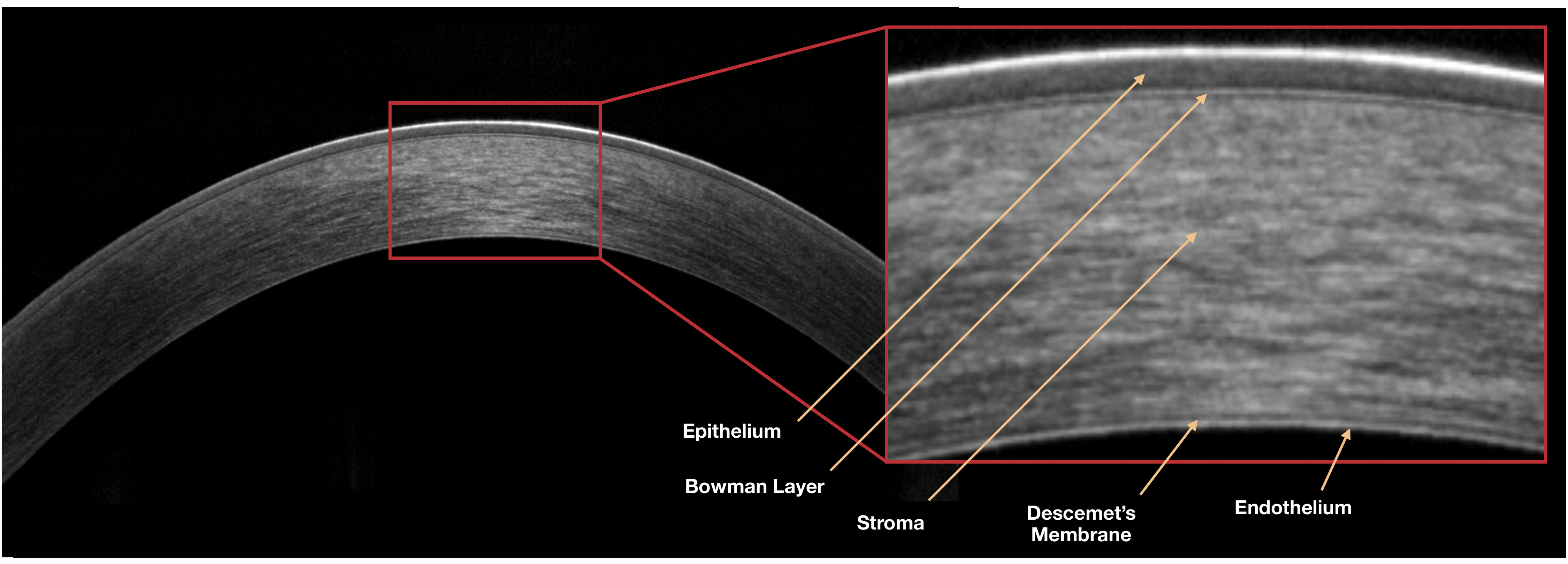}}
\caption{Representative clinical corneal OCT B-scan with an enlarged of view of the central corneal region. The principal cornea llayers are labelled to illustrate the anatomical interfaces targeted by the proposed boundary-localisation model.}
\label{fig:cornea_oct_scan_example}
\end{figure*}

Anterior segment optical coherence tomography (AS-OCT) has become an important imaging modality in corneal ophthalmology, offering non-invasive cross-sectional visualization of the cornea and its layered architecture. Based on low-coherence interferometry, AS-OCT provides micrometer-scale axial resolution and enables visualization of the major corneal layers, including the epithelium and stroma. In high-quality acquisitions, finer structures such as Bowman’s layer, Descemet’s membrane, and the endothelial interface may also be identified (see Fig.~\ref{fig:cornea_oct_scan_example}).

Beyond structural imaging, AS-OCT facilitates assessment of corneal morphology in several ocular conditions, including keratoconus, Fuchs endothelial corneal dystrophy (FECD), infectious keratitis, and post-surgical remodeling following procedures such as photorefractive keratectomy (PRK). Reliable delineation of corneal layers and their interfaces allows the extraction of clinically relevant measurements, including epithelial and stromal thickness maps, corneal geometry descriptors, and structural biomarkers that may support diagnosis, disease monitoring, and surgical planning~\cite{AranhadosSantos1,Werkmeister:17}.

Accurate localization of corneal interfaces remains challenging, however. These boundaries can appear faint, noisy, or locally ambiguous due to pathology-related deformation, speckle noise, acquisition artifacts, and device differences. Manual annotation is time-consuming, expert-dependent, and difficult to perform consistently, especially in large or longitudinal datasets. Automated segmentation methods are therefore needed to improve reproducibility, reduce analysis time, and support the development of objective imaging biomarkers for clinical and research applications.

Deep learning has increasingly been applied to corneal OCT layer and interface segmentation, building on convolutional and fully convolutional architectures for dense image prediction.
U-Net-based encoder--decoder~\cite{Ronneberger2015} models and related architectures have been used to delineate corneal layers and interfaces, with subsequent approaches introducing boundary-guided supervision, attention mechanisms, or multi-scale feature extraction to improve localization of thin or low-contrast structures~\cite{CorneaNet,Mathai2018,WANG2021108158,GARCIAMARIN2023106342}.

While these approaches have demonstrated accurate corneal layer segmentation within their respective datasets and acquisition settings, their evaluation has predominantly focused on relatively consistent imaging conditions, often using the same scanner or acquisition protocol.
Consequently, it remains unclear how well such models generalize to OCT systems that were entirely unseen during training, where differences in image resolution, contrast, noise characteristics, and scanner-specific appearance can introduce substantial domain shifts.
Broader clinical translation is further constrained by the limited availability of public datasets with detailed annotations of individual corneal layers, which complicates reproducible benchmarking and direct comparison between methods.

These challenges motivate segmentation methods that preserve local anatomical detail while maintaining coherent full-image boundary structure under different imaging conditions. In this work, we propose ARCOS, a patch-based framework for corneal boundary detection in AS-OCT images. The method processes overlapping full-resolution patches, predicts boundary heatmaps for corneal interfaces, and reconstructs full-resolution boundary estimates by combining patch-level predictions across the entire B-scan. This design avoids resizing the entire image to a fixed input resolution and accommodates variations in image dimensions, pixel spacing, field of view, and scanner-specific characteristics.

Rather than treating corneal layer segmentation as a conventional region classification problem, the proposed method defines the task as boundary localization. This formulation reflects the geometry of corneal OCT, where clinically relevant measurements depend on thin, continuous, and anatomically ordered interfaces rather than on large object-like regions. 

The method is designed to improve interface localization, reduce dependence on user-defined parameter tuning, and support segmentation across different devices and pathological cases.

\section{Materials and Methods}
\subsection{Devices and Datasets}

The datasets used in this study included clinical acquisitions and public datasets obtained with five AS-OCT systems from different manufacturers. Clinical data were acquired using three SD-OCT systems: RTVue-XR Avanti (Optovue Inc., Fremont, CA, USA), Solix (Visionix/Optovue Inc., Fremont, CA, USA), and MS-39 (Costruzione Strumenti Oftalmici, Firenze, Italy). Two additional SS-OCT public datasets, MCOA~\cite{mcoa}, acquired using a TowardPi device (TowardPi Medical Technology Co., Ltd., Beijing, China), and AIDK~\cite{aidk}, acquired using CASIA SS-1000 (Tomey Corporation, Nagoya, Japan), were included to further assess generalization to unseen acquisition technologies and image characteristics. Hereafter, these acquisition devices are referred to as Avanti, Solix, MS-39, TowardPi, and CASIA SS-1000.

The evaluated systems differed in OCT modality,  resolution, field of view, and pixel spacing, as summarised in Table~\ref{tab:oct_device_specs}. These differences affect both the physical sampling of corneal anatomy and the appearance of tissue interfaces, providing a heterogeneous setting for evaluating cross-device segmentation performance.

\begin{table}[!h]
\caption{
Technical characteristics of the OCT systems used in this study. 
Optical resolution and field of view (FOV) are reported from device specifications. 
Pixel spacing was estimated from image dimensions and scan size.
}
\label{tab:oct_device_specs}
\centering
\scriptsize
\renewcommand{\arraystretch}{1.05}
\setlength{\tabcolsep}{3pt}

\begin{adjustbox}{max width=\columnwidth}
\begin{tabularx}{\columnwidth}{
>{\raggedright\arraybackslash}p{70pt}
*{5}{>{\centering\arraybackslash}X}
}
\toprule
\textbf{Characteristic} &
\textbf{Avanti} &
\textbf{Solix} &
\textbf{MS-39} &
\textbf{TowardPi} &
\textbf{CASIA SS-1000} \\
\midrule

\rowcolor{gray!12}
\multicolumn{6}{l}{\textit{OCT technology}} \\
Type &
SD-OCT &
SD-OCT &
SD-OCT &
SS-OCT &
SS-OCT \\

\addlinespace[0.15em]

\rowcolor{gray!12}
\multicolumn{6}{l}{\textit{Nominal optical resolution}} \\
Axial resolution (\(\mu\mathrm{m}\)) &
5.0 &
5.0 &
3.6 &
1.9 &
10.0 \\

Lateral resolution (\(\mu\mathrm{m}\)) &
15.0 &
18.0 &
35.0 &
1.0 &
30.0 \\

\addlinespace[0.15em]

\rowcolor{gray!12}
\multicolumn{6}{l}{\textit{Scan geometry and image sampling}} \\
FOV (mm) &
\(8 \times 3\) &
\(8 \times 3\) &
\(16 \times 8\) &
\(24 \times 15\) &
\(16 \times 6\) \\

Axial pixel spacing (\(\mu\mathrm{m/px}\)) &
4.31 &
3.91 &
3.60 &
4.10 &
6.00 \\

Lateral pixel spacing (\(\mu\mathrm{m/px}\)) &
7.81 &
4.10 &
4.80 &
6.20 &
9.30 \\

\bottomrule
\end{tabularx}
\end{adjustbox}
\end{table}

In addition to acquisition-related variability, the datasets also differed in clinical composition. The final evaluation set included healthy corneas, post-surgical PRK cases, Fuchs' endothelial corneal dystrophy, and infectious keratitis, as reported in Table~\ref{tab:datasets_summary}. 
Together, this provides a diverse evaluation setting under both device-related and biological variability.
\begin{table}[!h]
\caption{
Summary of OCT datasets used in the study.
}
\label{tab:datasets_summary}
\centering
\footnotesize
\renewcommand{\arraystretch}{1.05}
\setlength{\tabcolsep}{3pt}

\begin{adjustbox}{max width=\columnwidth}
\begin{tabular}{
p{58pt}
p{64pt}
S[table-format=3.0,detect-weight=true]
S[table-format=3.0,detect-weight=true]
p{88pt}
}
\toprule

\textbf{Device} &
\textbf{Cohort} &
{\textbf{Subjects}} &
{\textbf{Scans}} &
\textbf{Notes} \\

\midrule

\rowcolor{gray!12}
\multicolumn{5}{l}{\textit{SD-OCT}} \\

Avanti &
Myopic PRK &
41 &
169 &
65\% pre-op, 35\% post-op. \\

Avanti &
Hyperopic PRK &
11 &
28 &
75\% pre-op, 25\% post-op. \\

Avanti &
Fuchs' dystrophy &
36 &
83 &
FECD. \\

Solix &
Healthy &
3 &
20 &
-- \\

MS-39 &
Healthy &
11 &
20 &
-- \\

\addlinespace[0.12em]

\rowcolor{gray!12}
\multicolumn{5}{l}{\textit{SS-OCT}} \\

TowardPi &
Healthy &
17 &
20 &
-- \\

CASIA SS-1000 &
Keratitis &
7 &
20 &
-- \\

\addlinespace[0.12em]

\rowcolor{gray!8}
\textbf{Total} &
\textbf{--} &
\textbf{126} &
\textbf{360} &
\textbf{--} \\

\bottomrule
\end{tabular}
\end{adjustbox}

\end{table}
\FloatBarrier
\subsection{Annotation Protocol}

Certain thin corneal interfaces approach the axial resolution limits of standard clinical OCT systems, making reliable delineation of all individual layers difficult in some scans. The annotation protocol was defined according to visually and clinically distinguishable interfaces. Five boundary lines were manually delineated for each B-scan:

\begin{enumerate}
    \item Boundary 1: Air--epithelium
    \item Boundary 2: Epithelium--epithelial basal layer + Bowman's layer
    \item Boundary 3: Epithelial basal layer + Bowman's layer--stroma
    \item Boundary 4: Stroma--Descemet membrane + endothelium
    \item Boundary 5: Descemet membrane + endothelium--anterior chamber
\end{enumerate}

Annotations were performed by a trained operator following a standardized protocol to ensure consistent delineation across scans and acquisition systems. These annotations were used as the reference ground truth for model training and evaluation.

To reduce annotation time and preserve spatial consistency, a sparse point-based strategy was used. For each boundary, the operator placed about 30 to 50 control points along the visible interface. These points were then used to reconstruct the complete boundary profile. For regular corneal geometries, a sixth-degree polynomial fit was used. Piecewise cubic spline interpolation was used in cases with local curvature changes or irregular boundary morphology. All fitted boundaries were visually inspected and corrected manually when necessary to ensure anatomical plausibility.

Annotation consistency was checked using 20 images from the myopic PRK and Fuchs' dystrophy cohorts. Three additional trained operators independently re-annotated these images using the same protocol. This analysis estimated intrinsic manual annotation variability and provided a reference level for automated segmentation performance.

As shown in Table~\ref{tab:interoperator_mae}, inter-operator disagreement was low, with a pooled mean absolute error of \(1.06 \pm 0.56\) pixels across all pairwise comparisons. Larger deviations were observed mainly near the lateral image regions, where signal quality decreases, and interface delineation becomes less reliable.

\begin{table}[!h]
\caption{
Pairwise inter-operator variability across all annotated boundaries. 
Values are reported as mean absolute error (MAE) between boundary positions. 
}
\label{tab:interoperator_mae}
\centering
\scriptsize
\renewcommand{\arraystretch}{1.05}
\setlength{\tabcolsep}{3pt}

\begin{adjustbox}{max width=\columnwidth}
\begin{tabular}{
p{46pt}
p{52pt}
S[table-format=1.2,detect-weight=true]
S[table-format=1.2,detect-weight=true]
S[table-format=1.2,detect-weight=true]
S[table-format=1.2,detect-weight=true]
S[table-format=1.2,detect-weight=true]
S[table-format=3.0,detect-weight=true]
}
\toprule

\textbf{Operator 1} &
\textbf{Operator 2} &
{\textbf{Mean}} &
{\textbf{Median}} &
{\textbf{SD}} &
{\textbf{P90}} &
{\textbf{P95}} &
{\textbf{\(n\)}} \\

\midrule

\rowcolor{gray!12}
\multicolumn{8}{l}{\textit{Reference Operator Comparisons}} \\

Reference & Operator A & 0.89 & 0.76 & 0.53 & 1.52 & 2.04 & 100 \\
Reference & Operator B & 0.79 & 0.70 & 0.39 & 1.29 & 1.59 & 100 \\
Reference & Operator C & 1.17 & 1.06 & 0.55 & 1.95 & 2.29 & 100 \\

\addlinespace[0.12em]

\rowcolor{gray!12}
\multicolumn{8}{l}{\textit{Inter-operator Comparisons}} \\

Operator A & Operator C & 1.22 & 1.13 & 0.61 & 2.15 & 2.37 & 100 \\
Operator A & Operator B & 1.01 & 0.89 & 0.51 & 1.64 & 1.86 & 100 \\
Operator B & Operator C & 1.31 & 1.25 & 0.55 & 1.93 & 2.34 & 100 \\

\addlinespace[0.12em]

\rowcolor{gray!8}
\textbf{All pairs} & \textbf{pooled} &
\textbf{1.06} &
\textbf{0.94} &
\textbf{0.56} &
\textbf{1.82} &
\textbf{2.18} &
\textbf{600} \\

\bottomrule
\end{tabular}
\end{adjustbox}

\end{table}
\subsection{Method Overview}

\begin{figure}[!t]
\centering
\includegraphics[width=\columnwidth]{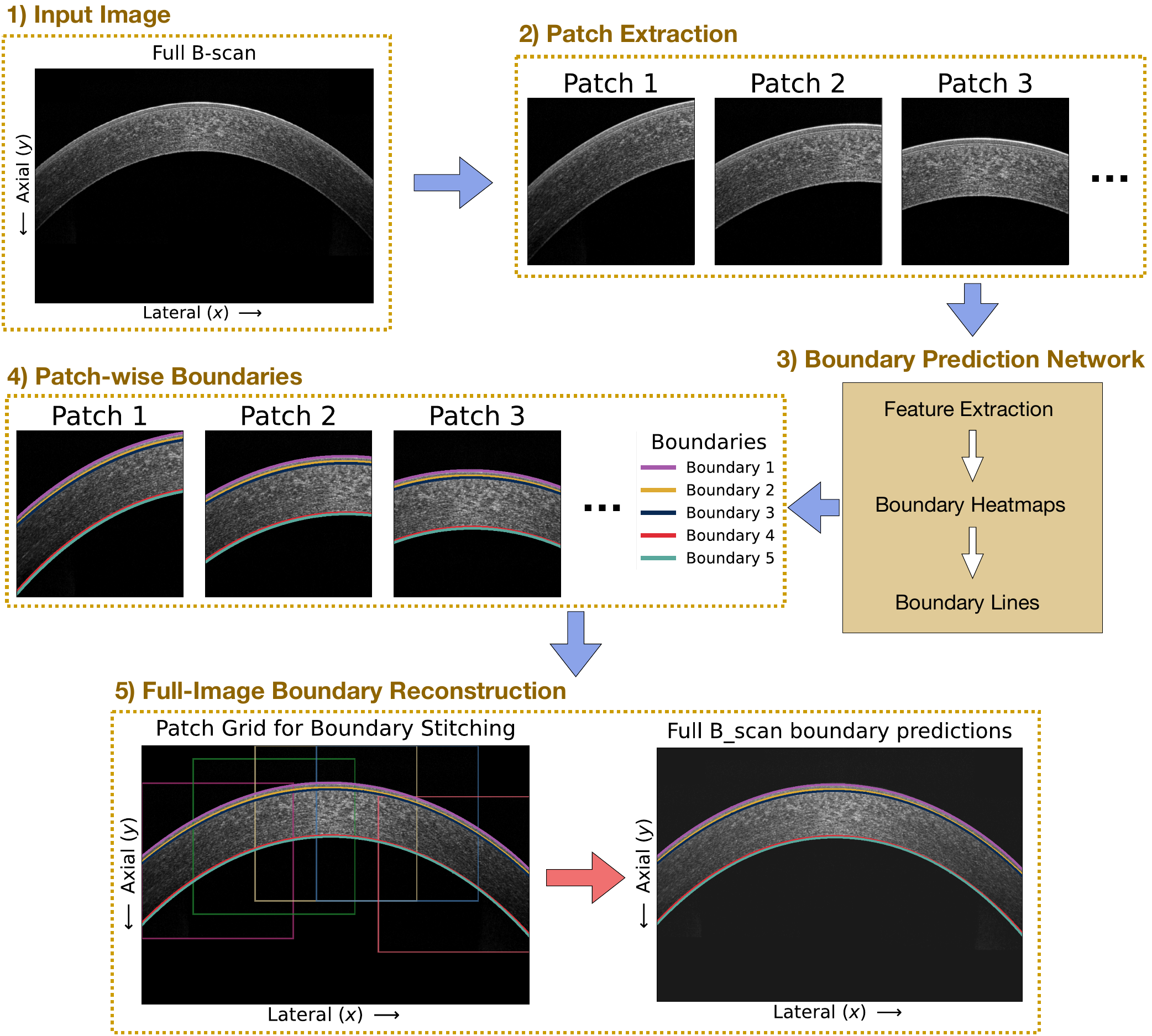}
\caption{
Overview of the proposed segmentation framework. 
An input AS-OCT B-scan is decomposed into overlapping patches, processed by the segmentation network to estimate boundary heatmaps, and reconstructed into full B-scan corneal interface predictions by merging the patch-level outputs.
}
\label{fig:method_overview}
\end{figure}

The proposed framework, illustrated in Fig.~\ref{fig:method_overview}, performs corneal boundary detection in AS-OCT B-scans by formulating the task as boundary localization rather than direct region classification. Given an input image, overlapping patches are extracted along the B-scan while preserving the native image resolution. 
Each patch is processed by a U-Net-based encoder--decoder network that predicts heatmaps for the five corneal interfaces. 
Patch-level predictions are then mapped back to their original image coordinates and fused using Gaussian-weighted stitching, which gives greater weight to the central region of each patch and attenuates edge-related artifacts. The reconstructed heatmaps are finally converted into continuous boundary profiles across the complete B-scan.

This design avoids resizing full B-scans to a fixed input size, reducing dependence on image dimensions, field of view, and scan geometry. Consequently, device-specific differences primarily affect the number and placement of extracted patches rather than the network input representation. The heatmap formulation reduces the sparsity of one-pixel boundary labels and provides spatial tolerance for thin structures and annotation uncertainty, while overlapping patches and weighted stitching reduce border artifacts during reconstruction.

\subsection{Patch Extraction}

A lightweight boundary detection procedure was first applied to estimate the approximate anterior and posterior corneal interfaces. 
The first prominent peak was identified as the anterior boundary, while the last prominent peak was used to estimate the posterior corneal interface. 
The detected surfaces were used exclusively to guide patch positioning and ensure consistent coverage of the corneal tissue region, rather than to provide precise anatomical localization.

Patches of $384 \times 384$ pixels were extracted with a target lateral overlap of 50\%; when exact overlap was not possible, the lateral stride was adjusted uniformly to ensure complete B-scan coverage.
Because neighboring regions are vertically displaced by the natural curvature of the cornea, the vertical position of each patch was adapted according to the estimated surfaces to maintain full visibility of the corneal tissue region.

The selected patch size was chosen to preserve the complete corneal thickness while providing sufficient lateral anatomical context for stable boundary estimation. 
For all OCT systems included in this study, a patch height of $384$ pixels was sufficient to fully contain the corneal regions within the extracted region. 
However, the model can be used with different patch sizes if necessary, as discussed later in section \ref{sec:model_robustness}.
\subsection{Network Architecture and Heatmap Prediction}

ARCOS formulates corneal layer delineation as a boundary heatmap regression task. The network predicts five boundary response maps, one for each anatomical interface. Manual annotations are converted into dense supervision targets by placing a Gaussian profile in each lateral column, centred at the annotated axial position of the corresponding interface.

The network architecture builds on a U-Net encoder--decoder design with multi-scale feature fusion and decoder refinement. A ConvNeXt-Small backbone~\cite{convnext} is used as the encoder to extract hierarchical feature representations. The decoder progressively restores spatial resolution by combining features across scales. Before fusion, attention gates~\cite{attentionunet} are applied to the skip connections to suppress irrelevant activations and emphasize boundary-relevant information.

The decoder produces a fused multi-scale representation, denoted by \(\mathbf{F}\), from which an initial set of boundary heatmaps \(\mathbf{H}_{\mathrm{init}}\) is predicted. These heatmaps are refined using a lightweight self-conditioned residual module based on feature-wise linear modulation (FiLM)~\cite{film}. A compact patch-level embedding is first obtained by global average pooling, followed by an MLP:
\begin{equation}
\mathbf{r} = \phi_{\mathrm{MLP}}(\mathrm{GAP}(\mathbf{F})).
\end{equation}
The refinement branch receives the concatenated representation \(\mathbf{Z}=[\mathbf{F},\mathbf{H}_{\mathrm{init}}]\), which is projected using a \(1\times1\) convolution, instance normalization, and ReLU activation to produce features \(\mathbf{U}\). FiLM parameters generated from \(\mathbf{r}\) modulate these features as
\begin{equation}
\widetilde{\mathbf{U}}_{c,:,:} = \gamma_c \mathbf{U}_{c,:,:} + \beta_c,
\quad
\boldsymbol{\gamma}=W_{\gamma}\mathbf{r}+b_{\gamma}, \quad
\boldsymbol{\beta}=W_{\beta}\mathbf{r}+b_{\beta}.
\end{equation}
A final \(1\times1\) convolution predicts a residual correction \(\Delta\mathbf{H}\), giving
\begin{equation}
\mathbf{H}_{\mathrm{main}} = \mathbf{H}_{\mathrm{init}} + \alpha \Delta\mathbf{H},
\end{equation}
where \(\alpha=0.4\) is fixed for all experiments. This residual formulation preserves the main prediction pathway while allowing patch-level appearance information to guide local heatmap refinement.

During inference, explicit boundary coordinates are obtained using a column-wise hard argmax applied independently to each predicted interface heatmap. This converts dense heatmap outputs into boundary profiles while avoiding the averaging effects associated with soft-coordinate extraction when responses are broad or locally uncertain.

\begin{figure}[!t]
\centerline{\includegraphics[width=\columnwidth]{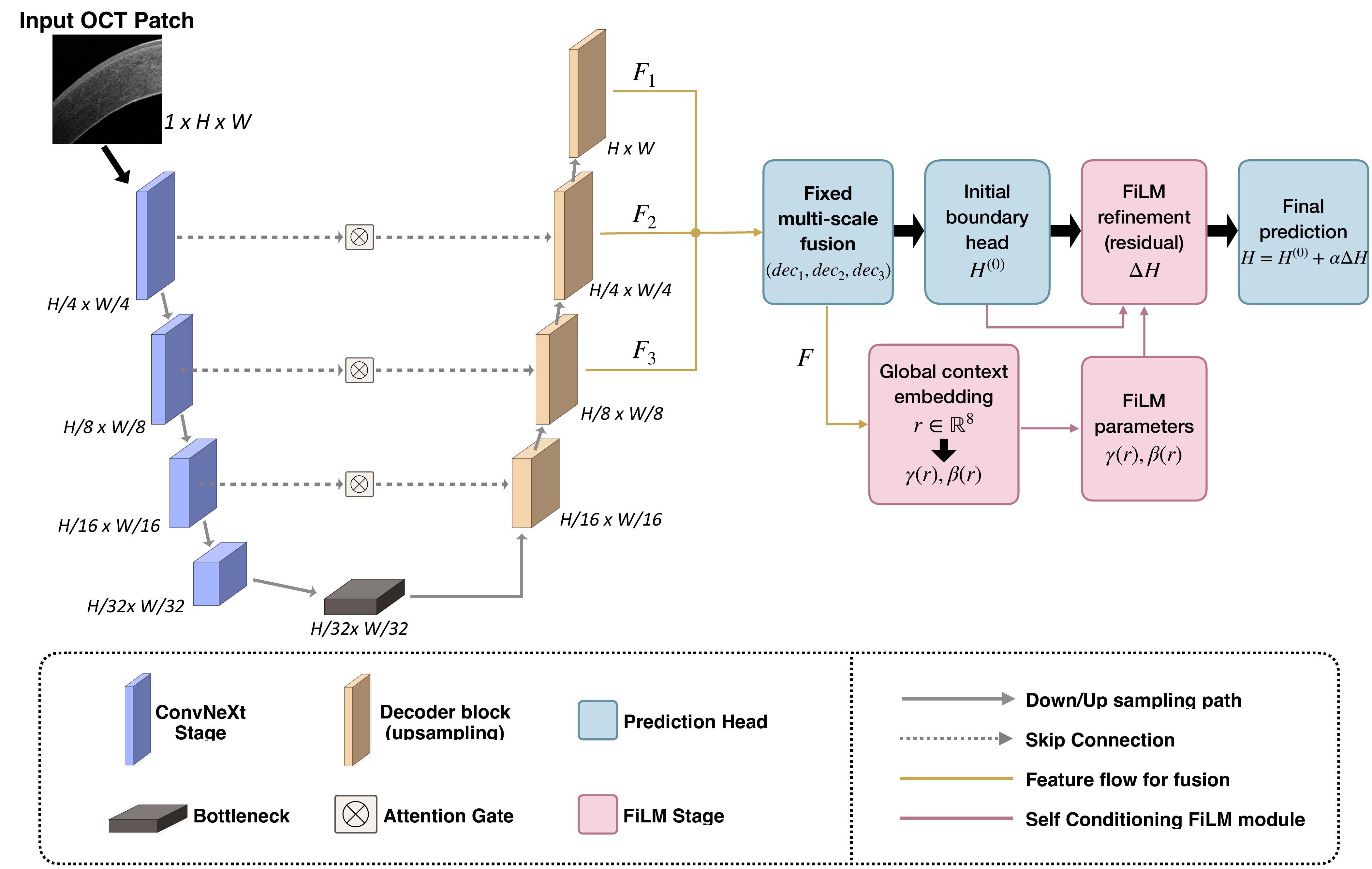}}
\caption{
The proposed network predicts boundary heatmaps. A ConvNeXt-Small encoder extracts hierarchical features. An attention-gated decoder combines these features to predict initial corneal interface heatmaps. A lightweight self-conditioned FiLM refinement module predicts residual heatmap corrections, using a patch-level embedding for conditioning.
}
\label{fig:architecture}
\end{figure}

An overview of the complete architecture is shown in Fig.~\ref{fig:architecture}.

\subsection{Training Strategy}

The model was trained using the \emph{Adaptive Wing} (AW) loss~\cite{awl}, which is well suited to heatmap regression. Compared with conventional losses such as mean squared error, AW loss adaptively balances errors around high-confidence target regions and background areas, encouraging accurate localization of boundary responses without allowing easy background pixels to dominate the optimization.

\begin{figure*}[!h]
\centerline{\includegraphics[width=\textwidth]{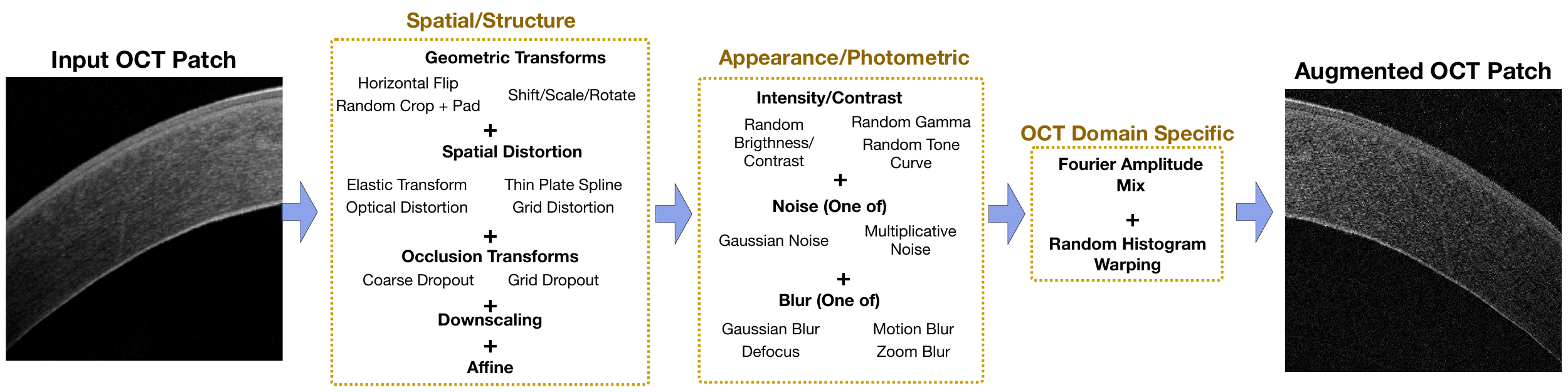}}
\caption{
Overview of the curriculum augmentation strategy used during training. 
Spatial, photometric, and OCT-specific frequency/intensity transformations are introduced progressively by increasing their probability or strength over training. 
This exposes the model to variability in corneal shape, acquisition geometry, contrast, speckle-like texture, and scanner-dependent image appearance while maintaining stable early optimisation.
}
\label{fig:augmentations}
\end{figure*}

To improve generalization across heterogeneous OCT imaging conditions, training used a multi-component augmentation strategy targeting three main sources of variability: acquisition geometry and corneal shape, image appearance and contrast, and OCT-specific frequency and intensity characteristics. These were addressed using spatial and structural transformations, photometric perturbations, and frequency- and intensity-domain modifications, respectively. Augmentations were applied using a curriculum schedule, in which the probability or strength of each transformation was increased progressively during training. This allowed the model to first learn stable boundary localization from moderately perturbed examples before being exposed to stronger appearance and acquisition variability. An overview of the augmentation workflow is shown in Fig.~\ref{fig:augmentations}.

In addition to standard geometric and photometric augmentations, two OCT-specific transformations were included. Inspired by Fourier-domain adaptation and Fourier-based domain generalization methods~\cite{Yang2020FDA,Xu2023AmpMix}, Fourier amplitude perturbation was applied by decomposing each image into magnitude and phase components in the Fourier domain, adding Gaussian noise to the magnitude spectrum, and reconstructing the image using the original phase. This preserves the main spatial organization of the corneal structures while introducing variations in frequency content and texture. Unlike target-domain adaptation methods, the transformation does not require access to target-device images and is used only to increase the range of scanner-like appearance variations observed during training.  As a second OCT-specific transformation, histogram warping was applied to smooth the intensity distribution, further modeling variations in contrast, brightness response, and scanner-specific image rendering.

Together, these augmentations expose the model to variations in shape, sampling, contrast, texture, and intensity distribution that may occur across OCT systems and acquisition protocols. This training strategy complements the patch-based heatmap formulation by improving robustness to appearance changes that are not fully captured by standard augmentations alone.
\section{Experiments}
\subsection{Data Splits}

Avanti was used as the training device, while all other OCT systems were reserved exclusively for inference-time evaluation. The Solix, MS-39, TowardPi, and CASIA SS-1000 datasets were not used for parameter selection and were used solely to assess method flexibility and cross-device generalization.

Within the Avanti device, the myopic PRK cohort and 90\% of the Fuchs' dystrophy cohort were used for training and validation, with an 80/20 split. The in-distribution Avanti test set consisted of the hyperopic PRK cohort and the remaining 10\% of Fuchs' dystrophy cases. This provided a held-out test set containing both post-surgical and pathological cases.

Patient-level separation was enforced in all splits: all scans from a given patient were assigned to the same training, validation, or test partition to avoid data leakage. The resulting effective split was approximately 70\% training, 18\% validation, and 12\% testing.

\subsection{Preprocessing and Standardization}

Despite substantial differences across OCT systems, all datasets were processed using a common standardization pipeline, without device-specific preprocessing or post-processing. Intensity preprocessing was limited to linear scaling to preserve the native appearance characteristics of each acquisition system.

For all datasets, \(384 \times 384\)-pixel patches were extracted directly from the original B-scans. Consequently, device differences were reflected in the number of extracted patches per scan and in the lateral overlap between neighboring patches, which ranged from approximately 50\% to 58\%.

Several devices showed reduced boundary visibility away from the corneal apex, particularly for thinner internal interfaces, leading to greater annotation uncertainty outside the central region. To ensure consistent evaluation across systems, metrics were computed within a common central 6\,mm region of interest (ROI), reducing the influence of peripheral signal degradation and annotation uncertainty. This effect was most pronounced in some SS-OCT scans, particularly from the CASIA SS-1000.

\subsection{Baseline Models}

The proposed framework was compared with representative baselines covering ablated, full-image, and previously proposed segmentation formulations.

\begin{itemize}
    \item \textbf{Basic Model}: an ablated version without multi-scale feature fusion, FiLM-based refinement, or OCT-specific augmentations.

    \item \textbf{Full Size (Original Size)}: the same architecture and training strategy as the proposed framework, but applied directly to full-width OCT B-scans instead of overlapping patches.

     \item \textbf{CorneaNet-based model}: a U-Net-based corneal OCT segmentation baseline reimplemented following CorneaNet~\cite{CorneaNet}, using region-based segmentation.

    \item \textbf{ScLNet-based Model}: a hybrid architecture following ScLNet~\cite{ScLNet}, combining boundary prediction with region-based segmentation supervision, operating on resized full-image inputs.
\end{itemize}

All baselines were trained and evaluated using the same dataset splits, preprocessing pipeline, and evaluation metrics. For models producing region-based segmentation maps rather than explicit boundary heatmaps, boundary coordinates were extracted from the predicted masks as the first pixel of the corresponding layer in each image column.
\subsection{Evaluation Metrics}

Segmentation performance was evaluated with boundary-based metrics after patch stitching and full-image boundary reconstruction. Unless otherwise stated, metrics were first computed per image and then averaged across scans from each OCT system to obtain device-level results.

The primary metric was \textbf{off-by-one accuracy}, hereafter referred to as accuracy, defined as the proportion of predicted boundary locations within \(\pm1\) pixel of the reference annotation. This tolerance accounts for small annotation uncertainty at weak or ambiguous interfaces. Mean absolute error (MAE) was used to quantify average boundary deviation in pixels. To assess geometric consistency and failure cases, Hausdorff distance (HD) and its 95th percentile variant (HD95) were also computed, with HD95 reducing the influence of isolated outliers.

Statistical comparisons were performed separately for each OCT device using image-level paired observations, thereby avoiding the masking of acquisition-specific effects. Significance was assessed using the paired Wilcoxon signed-rank test, selected due to the skewed distributions and localized outliers often observed in boundary error metrics. \(p\)-values were adjusted for multiple comparisons using the Benjamini--Hochberg false discovery rate correction, with adjusted \(p < 0.05\) considered significant.
\subsection{Implementation Details}

The proposed framework and its ablated variants were trained using the AdamW optimiser with an initial learning rate of \(10^{-3}\), a weight decay of \(10^{-4}\), and a ReduceLROnPlateau learning-rate scheduler. Boundary heatmap targets were generated using Gaussian profiles with \(\sigma = 2.5\) pixels. This value was selected empirically after evaluating alternative kernel widths, as it provided the best overall localisation performance. The curriculum augmentation schedule described earlier was applied during training, with augmentation strength increased linearly until reaching its maximum value at 60\% of the training process.

Models were trained with a batch size of 12 for up to 150 epochs. The CorneaNet-based and ScLNet-based baselines were trained using the optimization settings reported in their original publications. 

All experiments were implemented in PyTorch and performed on an NVIDIA Tesla V100 GPU.
\section{Results and Discussion}
\subsection{Training device results}

In-distribution performance is shown in Table~\ref{tab:baseline_comparison}, where the proposed model achieved the best overall localization performance, with an accuracy of 95.1\%, an MAE of 0.514 pixels, and an HD95 of 1.22 pixels. The Basic Model was the closest competitor, showing only a small performance gap, as expected under matched acquisition conditions where robustness-oriented components are less likely to show their full benefit.

\begin{table}[!h]
\caption{
Quantitative segmentation performance under in-distribution conditions. 
Metrics are reported as mean values over the test set. 
Best-performing results for each metric are highlighted in bold.
}
\label{tab:baseline_comparison}
\centering
\footnotesize
\renewcommand{\arraystretch}{0.85}
\setlength{\tabcolsep}{4pt}

\begin{adjustbox}{max width=\columnwidth}
\begin{tabular}{p{85pt}cccc}
\toprule
\textbf{Model} &
\textbf{Accuracy $\uparrow$} &
\textbf{MAE $\downarrow$} &
\textbf{HD95 $\downarrow$} &
\textbf{HD $\downarrow$} \\
\midrule

Proposed Model & \textbf{95.1} & \textbf{0.51} & \textbf{1.22} & 3.81 \\

\addlinespace[0.12em]

\rowcolor{gray!12}
\multicolumn{5}{l}{\textit{Baseline Models}} \\

Basic Model & 94.5 & 0.54 & 1.25 & 4.22 \\
Full Size & 84.8 & 0.80 & 1.75 & 3.19 \\
CorneaNet & 89.4 & 0.77 & 2.04 & 7.76 \\
ScLNet & 87.2 & 0.76 & 1.53 & \textbf{2.99} \\

\bottomrule
\end{tabular}
\end{adjustbox}

\end{table}

\begin{figure}[!h]
\centerline{\includegraphics[width=\columnwidth]{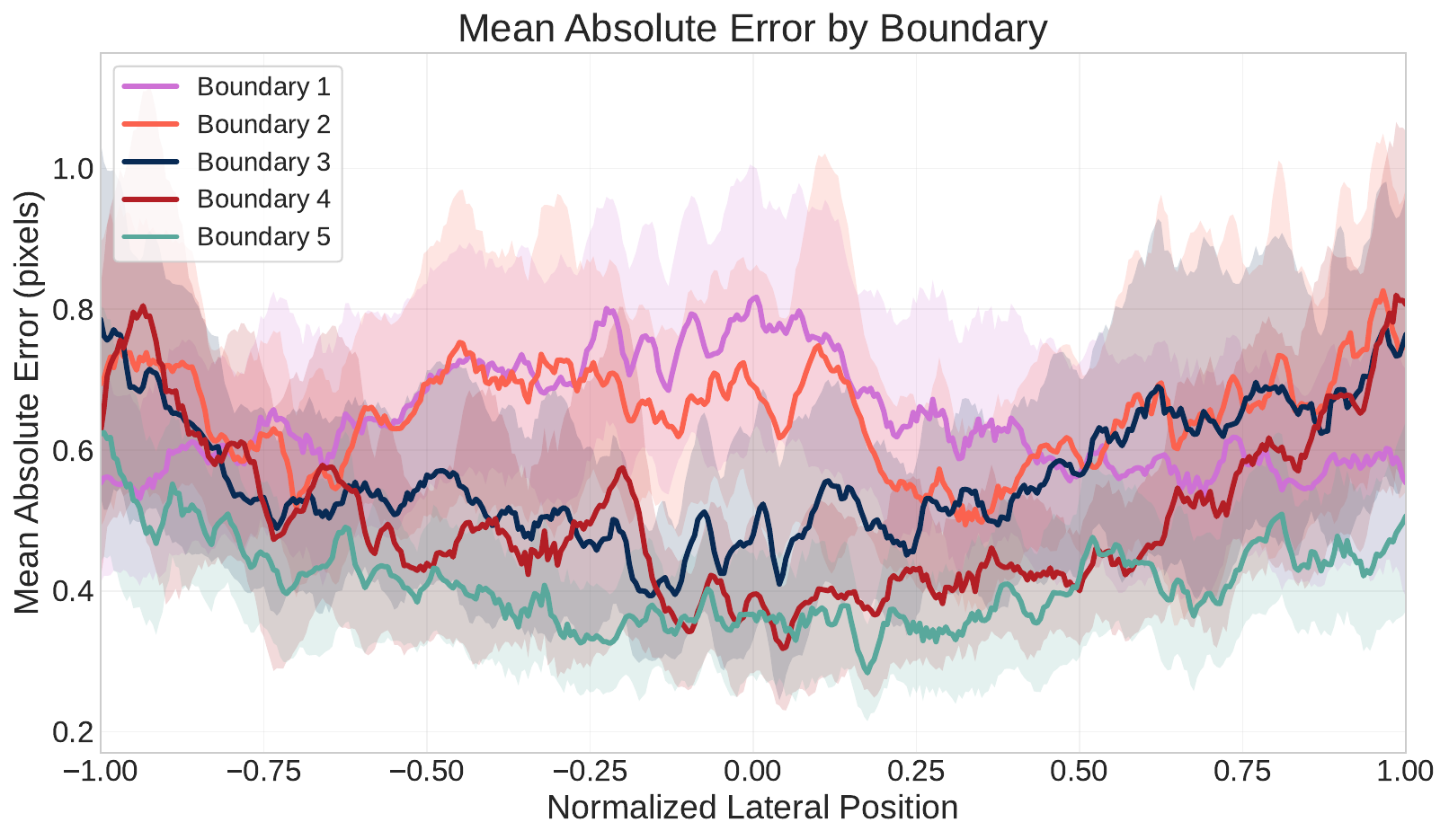}}
\caption{
Lateral variation of mean absolute error (MAE) for each corneal boundary. Shaded regions indicate confidence intervals across the test set.
}
\label{fig:mae_variation}
\end{figure}

Figure~\ref{fig:mae_variation} shows the lateral MAE distribution for each corneal boundary using the proposed model. Errors remained low centrally and increased moderately towards the periphery, where OCT signal quality and boundary visibility are often reduced. 
A small local increase was observed near the corneal apex for the most anterior interfaces, possibly reflecting stronger superficial signal and local ambiguity at the air--epithelium and epithelium--Bowman's boundaries. Nevertheless, MAE remained below 1 pixel for all boundaries, indicating stable localization across the evaluated field of view.

\begin{figure*}[!h]
\centerline{\includegraphics[width=\textwidth]{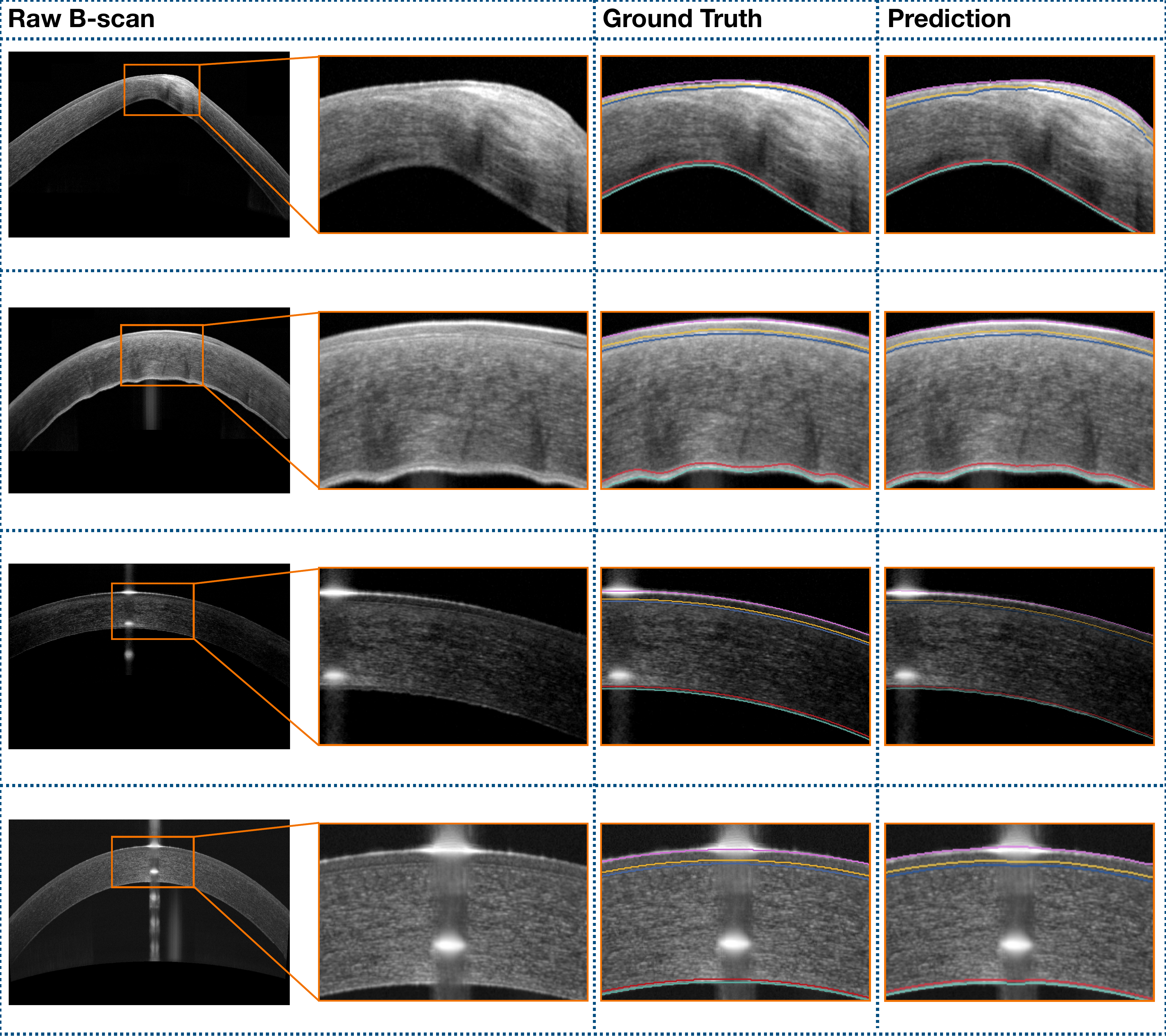}}
\caption{
Representative challenging cases, including keratoconus, FECD, low-signal acquisition, and central imaging artifacts. Ground-truth annotations and predicted boundaries are shown with magnified regions highlighting structural deformation and reduced image quality.
}
\label{fig:challenging_cases} 
\end{figure*}

Representative outputs in challenging cases are shown in Fig.~\ref{fig:challenging_cases}, including keratoconus, FECD, low-signal acquisitions, and central imaging artifacts. The predicted boundaries remained smooth and anatomically ordered despite local anatomical deformation and reduced boundary contrast. In particular, ARCOS correctly delineated boundaries in scans with central corneal artifacts, despite not being trained on images containing this artifact.

\subsection{Cross-Device Generalization}
\label{sec:cross_device_results}

Cross-device generalization was assessed by applying each model to OCT systems excluded from training. As shown in Table~\ref{tab:cross_device_results}, the proposed model demonstrated the most consistent performance across devices, achieving the highest accuracy and lowest MAE on all five systems, as well as the lowest HD and HD95 on four of them. The largest gains were observed for MS-39, TowardPi, and CASIA SS-1000, where baseline models exhibited more pronounced degradation in localization accuracy and Hausdorff-based metrics.
\begin{table}[h]
\caption{
Cross-device segmentation performance across OCT acquisition systems. Best values are highlighted in bold.
}
\label{tab:cross_device_results}
\centering
\footnotesize
\renewcommand{\arraystretch}{0.85}
\setlength{\tabcolsep}{4pt}

\begin{adjustbox}{max width=\columnwidth}
\begin{tabular}{p{50pt}ccccc}
\toprule
\textbf{Model} &
\textbf{Avanti} &
\textbf{Solix} &
\textbf{MS39} &
\textbf{TowardPi} &
\textbf{CASIA SS-1000} \\
\midrule

\rowcolor{gray!12}
\multicolumn{6}{l}{\textit{Accuracy $\uparrow$}} \\
Proposed & \textbf{95.1} & \textbf{94.8} & \textbf{86.1} & \textbf{75.9} & \textbf{80.5} \\
Basic & 94.5 & 93.7 & 68.3 & 67.9 & 76.4 \\
Full Image & 84.8 & 66.6 & 68.3 & 56.6 & 51.7 \\
CorneaNet & 89.4 & 87.6 & 53.7 & 62.5 & 68.9 \\
ScLNet & 87.2 & 83.1 & 76.6 & 37.7 & 79.1 \\

\addlinespace[0.12em]

\rowcolor{gray!12}
\multicolumn{6}{l}{\textit{MAE $\downarrow$}} \\
Proposed & \textbf{0.51} & 0.67 & \textbf{0.79} & \textbf{1.01} & \textbf{0.95} \\
Basic & 0.54 & \textbf{0.61} & 1.35 & 1.23 & 1.04 \\
Full Image & 0.80 & 1.98 & 1.31 & 1.97 & 3.16 \\
CorneaNet & 0.77 & 0.91 & 33.1 & 3.66 & 2.57 \\
ScLNet & 0.76 & 5.25 & 1.11 & 6.41 & 0.97 \\

\addlinespace[0.12em]

\rowcolor{gray!12}
\multicolumn{6}{l}{\textit{HD95 $\downarrow$}} \\
Proposed & \textbf{1.22} & \textbf{1.24} & \textbf{1.88} & \textbf{2.05} & 2.11 \\
Basic & 1.25 & 1.32 & 2.58 & 2.32 & 2.20 \\
Full Image & 1.75 & 6.02 & 2.65 & 4.19 & 8.49 \\
CorneaNet & 2.04 & 2.75 & 5.51 & 4.42 & 8.17 \\
ScLNet & 1.53 & 18.4 & 2.30 & 24.4 & \textbf{1.95} \\

\addlinespace[0.12em]

\rowcolor{gray!12}
\multicolumn{6}{l}{\textit{HD $\downarrow$}} \\
Proposed & \textbf{3.81} & \textbf{3.48} & \textbf{4.15} & \textbf{2.79} & 3.77 \\
Basic & 4.22 & 4.68 & 13.7 & 4.04 & \textbf{3.41} \\
Full Image & 3.19 & 31.4 & 6.55 & 29.8 & 20.4 \\
CorneaNet & 7.76 & 17.1 & 48.5 & 12.7 & 47.9 \\
ScLNet & 2.99 & 38.9 & 16.6 & 57.4 & 4.62 \\

\bottomrule
\end{tabular}
\end{adjustbox}
\end{table}

\begin{figure}[h]
\centering
\includegraphics[width=\columnwidth]{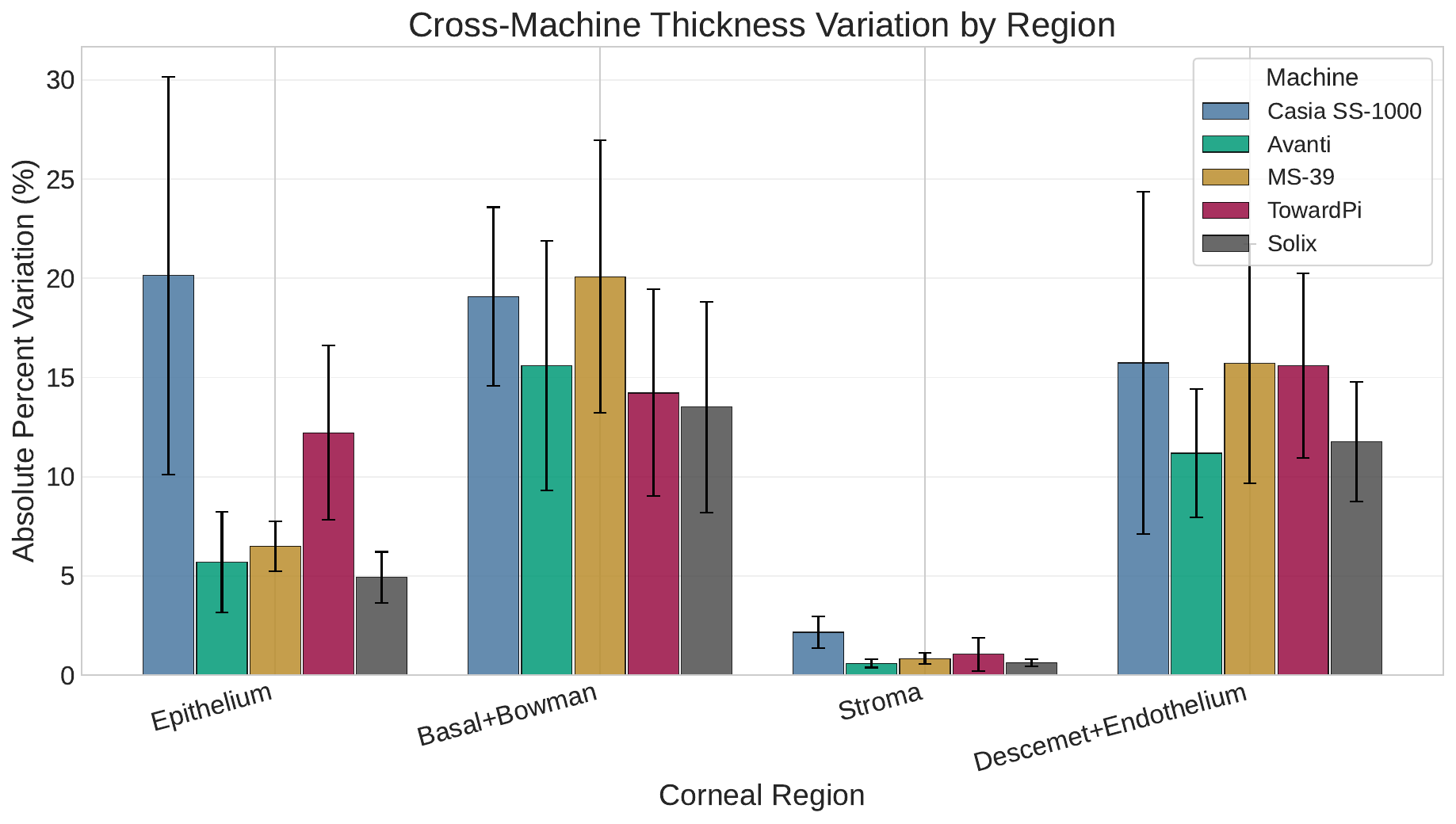}
\caption{
Relative thickness error of ARCOS across corneal regions and OCT acquisition systems. Percentage errors are larger in thinner layers, where small localization differences have a greater proportional effect on thickness estimation.
}
\label{fig:cross_device_thickness_variation}
\end{figure}

Figure~\ref{fig:cross_device_thickness_variation} shows relative thickness error across corneal regions and acquisition systems. As expected, the largest percentage deviations occurred in thinner layers, where small boundary errors have a greater proportional effect on thickness estimates. Thickness errors nevertheless remained controlled across devices, supporting quantitative pachymetric analysis under heterogeneous imaging conditions.

\begin{figure*}[!h]
\centering
\includegraphics[width=\textwidth]{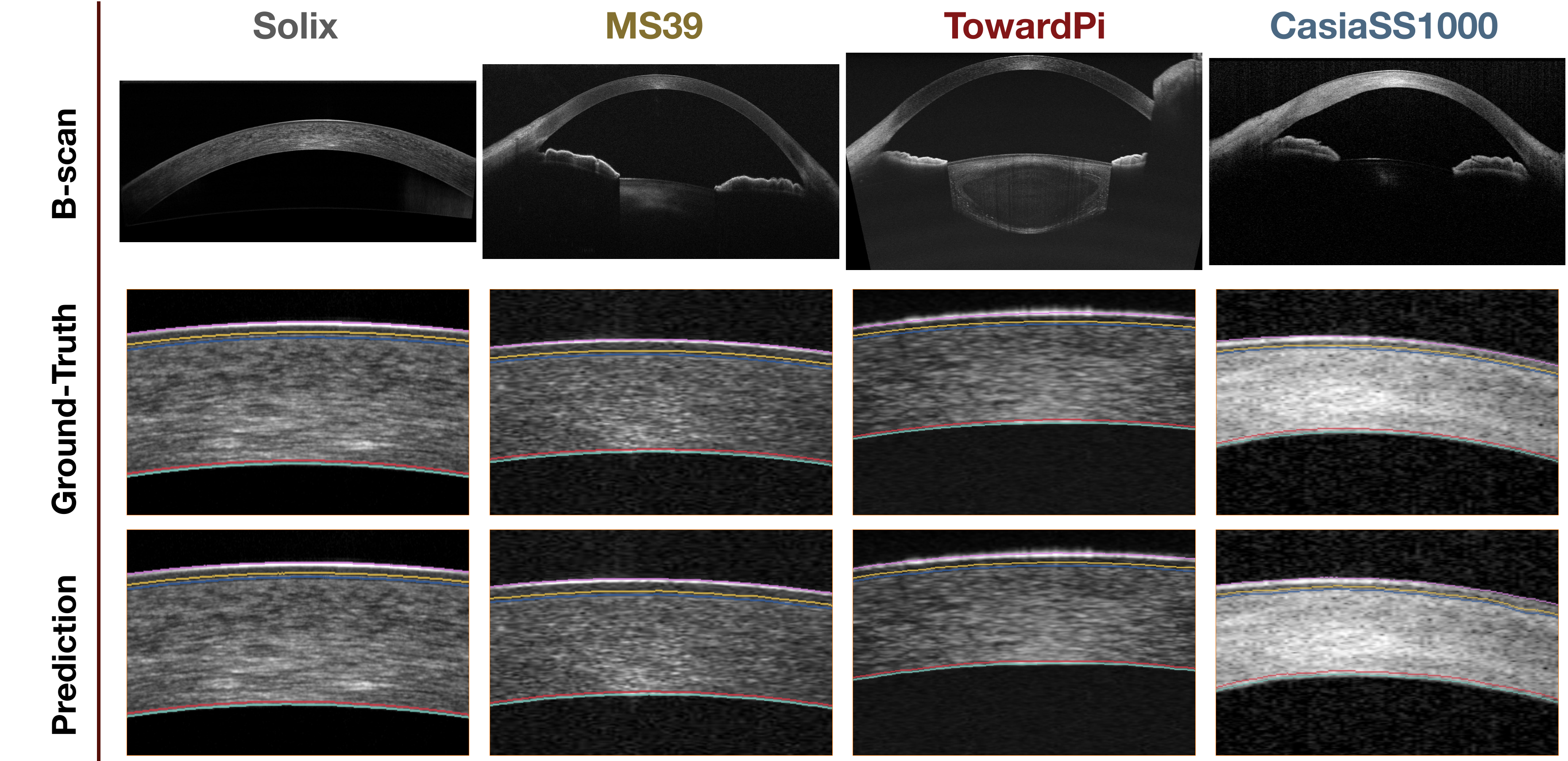}
\caption{
Representative cross-device segmentation results on OCT images acquired with different systems. Ground-truth annotations and predicted boundaries are shown for each example, illustrating the preservation of continuous and anatomically ordered corneal interfaces under device-related domain shift.
}
\label{fig:other_oct_cases}
\end{figure*}

Representative qualitative results are shown in Fig.~\ref{fig:other_oct_cases}. Despite visible differences in image appearance between devices, the predicted interfaces remain continuous and anatomically ordered in the selected examples. This qualitative behavior is consistent with the quantitative results and indicates that the proposed model preserves coherent corneal layer geometry across diverse acquisition conditions.

Overall, the combination of native-resolution patch processing and domain-oriented training improves stability under device shift, reducing the severe performance degradation observed in several baseline methods.

\subsection{ Model Robustness}
\label{sec:model_robustness}

To evaluate the effect of patch-scale spatial context, ARCOS was additionally tested with different inference patch sizes across all acquisition devices.

When inference was performed with larger patches (\(512 \times 512\)), performance remained close to the reference \(384 \times 384\) setting (Table~\ref{tab:patch_size_robustness}). This suggests that ARCOS is relatively stable under moderate increases in patch-scale spatial context. The result is consistent with the scale-related augmentations used during training, including cropping, zooming, and rescaling, which exposed the model to variations in effective field of view.

In contrast, performance decreased with smaller patches (\(256 \times 256\)). This suggests that, while the model is robust to larger contextual windows, very small patches may remove relevant corneal structure needed for accurate localization.
\begin{table}[h]
\caption{
Performance comparison between inference patch sizes across acquisition devices.
The model was trained using a reference patch size of \(384 \times 384\).
Values are reported separately for each device.
}
\label{tab:patch_size_robustness}
\centering
\normalsize
\renewcommand{\arraystretch}{0.95}
\setlength{\tabcolsep}{2.5pt}

\begin{adjustbox}{max width=\columnwidth}
\begin{tabular}{p{52pt}cccccccccccc}
\toprule

& \multicolumn{3}{c}{\textbf{Accuracy $\uparrow$}} 
& \multicolumn{3}{c}{\textbf{MAE $\downarrow$}}
& \multicolumn{3}{c}{\textbf{HD95 $\downarrow$}}
& \multicolumn{3}{c}{\textbf{HD $\downarrow$}} \\

\cmidrule(lr){2-4}
\cmidrule(lr){5-7}
\cmidrule(lr){8-10}
\cmidrule(lr){11-13}

\textbf{Device}
& \textbf{256}
& \textbf{384}
& \textbf{512}
& \textbf{256}
& \textbf{384}
& \textbf{512}
& \textbf{256}
& \textbf{384}
& \textbf{512}
& \textbf{256}
& \textbf{384}
& \textbf{512} \\

\midrule

Avanti
& 94.9 & 95.1 & 94.8
& 0.53 & 0.51 & 0.53
& 3.35 & 1.22 & 1.29
& 49.2 & 3.81 & 1.79 \\

Solix
& 93.9 & 94.8 & 94.4
& 0.96 & 0.67 & 0.58
& 4.36 & 1.24 & 1.27
& 31.15 & 3.48 & 2.51 \\

MS-39
& 76.1 & 86.1 & 84.4
& 3.94 & 0.79 & 0.86
& 18.4 & 1.88 & 1.97
& 83.9 & 4.15 & 5.59 \\

TowardPi
& 73.0 & 75.9 & 71.9
& 2.07 & 1.01 & 1.16
& 4.83 & 2.05 & 2.25
& 76.3 & 2.79 & 10.8 \\

CASIA SS-1000
& 76.8 & 80.5 & 81.1
& 1.05 & 0.95 & 0.91
& 2.34 & 2.11 & 2.01
& 3.17 & 3.77 & 2.89 \\

\addlinespace[0.12em]

\rowcolor{gray!12}
\textbf{Mean}
& 82.9 & \textbf{86.5} & 85.3
& 1.71 & \textbf{0.79} & 0.81
& 6.66 & \textbf{1.70} & 1.76
& 48.7 & \textbf{3.60} & 4.72 \\

\bottomrule
\end{tabular}
\end{adjustbox}

\end{table}
\subsection{Ablation}
\label{sec:ablation_patch}
To assess the contribution of individual components, ablation experiments were performed across all evaluation devices. Table~\ref{tab:ablation_results} compares the complete model with variants obtained by removing selected architectural and training components.

\begin{table}[!h]
\caption{
Ablation study across all evaluation devices. 
Values are reported as the mean across devices. 
Best-performing results for each metric are highlighted in bold.
}
\label{tab:ablation_results}
\centering
\footnotesize
\renewcommand{\arraystretch}{0.85}
\setlength{\tabcolsep}{4pt}

\begin{adjustbox}{max width=\columnwidth}
\begin{tabular}{
>{\raggedright\arraybackslash}p{105pt}
cccc
}
\toprule
\textbf{Model Variant} &
\textbf{Accuracy $\uparrow$} &
\textbf{MAE $\downarrow$} &
\textbf{HD95 $\downarrow$} &
\textbf{HD $\downarrow$} \\
\midrule

Proposed Model & \textbf{86.5} & \textbf{0.79} & \textbf{1.70} & 3.60 \\

\addlinespace[0.12em]

\rowcolor{gray!12}
\multicolumn{5}{l}{\textit{Backbone}} \\

ConvNeXt Tiny & 84.0 & 0.82 & 1.76 & \textbf{3.25} \\
ResNet34 & 73.1 & 1.18 & 3.16 & 8.27 \\
EfficientNet-B0 & 77.7 & 0.99 & 1.99 & 4.70 \\

\addlinespace[0.12em]

\rowcolor{gray!12}
\multicolumn{5}{l}{\textit{Loss Function}} \\

Heatmap MSE & 76.8 & 1.48 & 3.63 & 24.3 \\

\addlinespace[0.12em]

\rowcolor{gray!12}
\multicolumn{5}{l}{\textit{Attention and FiLM Modules}} \\

No Attention & 83.7 & 0.84 & 1.77 & 3.55 \\
No FiLM & 84.2 & 0.83 & 1.82 & 3.81 \\
No Attention + No FiLM & 80.6 & 0.95 & 2.07 & 5.40 \\

\addlinespace[0.12em]

\rowcolor{gray!12}
\multicolumn{5}{l}{\textit{Multi-scale Fusion}} \\

No Fusion & 82.8 & 0.89 & 1.87 & 4.96 \\
No FiLM + No Fusion & 84.7 & 0.83 & 1.81 & 5.08 \\
No Attention + No Fusion & 84.4 & 0.84 & 1.81 & 5.24 \\

\addlinespace[0.12em]

\rowcolor{gray!12}
\multicolumn{5}{l}{\textit{OCT-specific Augmentation}} \\

No OCT Augmentation & 83.1 & 0.88 & 1.83 & 5.44 \\
No Attention + No OCT & 83.7 & 0.83 & 1.78 & 3.98 \\
No Fusion + No OCT & 77.3 & 1.08 & 2.13 & 6.41 \\
No FiLM + No OCT & 83.5 & 0.86 & 1.83 & 4.98 \\

\addlinespace[0.12em]

\rowcolor{gray!12}
\multicolumn{5}{l}{\textit{Combined Ablations}} \\

No FiLM + No Fusion + No OCT & 84.7 & 0.82 & 1.83 & 5.27 \\
No Attention + No FiLM + No Fusion & 79.4 & 1.0 & 1.92 & 5.24 \\
No Attention + No FiLM + No OCT & 79.8 & 0.96 & 1.96 & 6.09 \\
No Attention + No Fusion + No OCT & 79.8 & 0.96 & 1.96 & 4.75 \\
No Attention + No FiLM + No Fusion + No OCT & 80.2 & 0.96 & 1.94 & 6.01 \\

\bottomrule
\end{tabular}
\end{adjustbox}

\end{table}

The full model achieved the strongest overall performance, with the best balance between localization accuracy and geometric stability. The largest degradation was observed when Adaptive Wing heatmap supervision was replaced by mean squared error, increasing MAE from 0.79 to 1.48 pixels and markedly worsening HD95 and HD. This indicates that the heatmap loss is important for stable boundary localization.

Removing attention gates, FiLM-based refinement, or multi-scale fusion also reduced performance, with larger drops when components were removed jointly. Statistical testing confirmed that the full model significantly outperformed each ablated variant across the reported metrics after multiple-comparison correction, except for HD95 and HD when compared with the ConvNeXt-Tiny and no-attention ablations.

Overall, the ablation results indicate that the proposed performance gains arise from the combined effect of heatmap supervision, architectural refinement, multi-scale fusion, and OCT-specific augmentation, rather than from a single isolated component.

\subsection{Failure Cases}

\begin{figure*}[!h]
\centerline{\includegraphics[width=\textwidth]{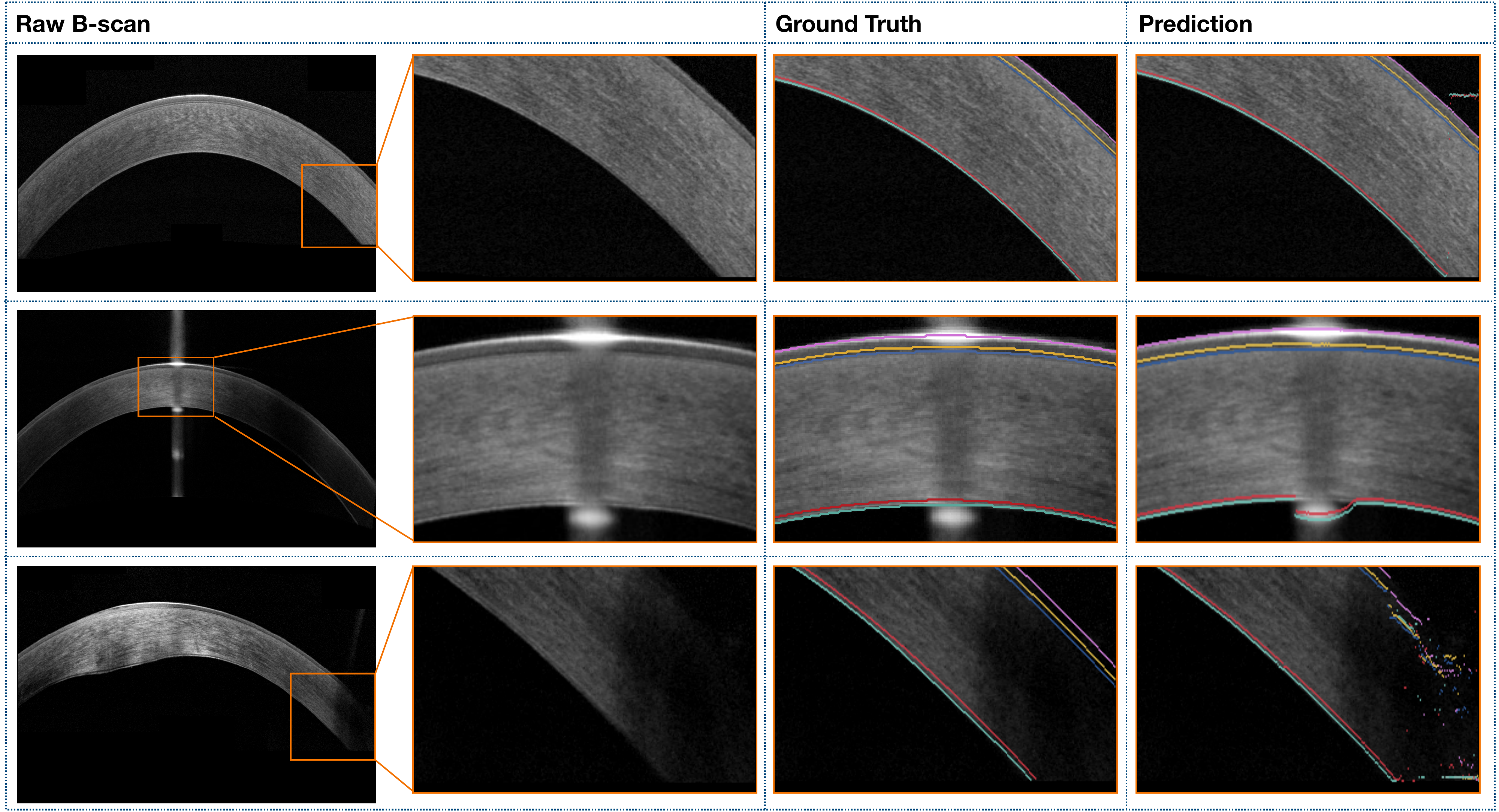}}
\caption{
Representative failure cases of the proposed model. Examples include peripheral patches where anatomical interfaces leave the visible image region, and scans affected by strong central artifacts that reduce visibility of the posterior boundary.
}
\label{fig:failure_cases}
\end{figure*}

Despite the strong overall performance across devices and challenging imaging conditions, several recurrent failure modes were identified (Fig.~\ref{fig:failure_cases}).

The most common failure mode occurred in edge patches, where one or more anatomical interfaces left the visible image region before the patch boundary was reached. In these cases, the model occasionally continued the predicted interface beyond the region where the corresponding structure was physically visible. 

The practical impact of this failure mode is limited for the main quantitative analyses in this study, since evaluation and thickness measurements were restricted to the central 6\,mm region of interest. Nevertheless, this issue could be mitigated by anatomical post-processing, for example, by masking predictions after an interface reaches the upper or lower image boundary.

A second failure mode was observed in scans affected by strong central imaging artifacts, where posterior interface localization could become unreliable within the corrupted region. For downstream analyses, these regions could be excluded or interpolated from neighboring valid predictions when appropriate to reduce the influence of localized errors.

Overall, these failure modes were relatively uncommon, occurring in approximately 7\% of test cases. They primarily occur in scan edges or reflect situations with strong imaging artifacts. These observations identify practical limitations of the current framework and suggest future improvements through uncertainty estimation, artifact detection, and anatomically constrained post-processing.

\subsection{Running Time}

Inference time was measured for the complete processing pipeline, including patch extraction, network prediction, heatmap reconstruction, and boundary stitching. The proposed method required, on average, 0.52\,s per B-scan on an NVIDIA Quadro P2000 GPU and approximately 3.8\,s per B-scan on an Intel Xeon These measurements indicate that full-scan segmentation can be performed within a few seconds on standard hardware, supporting practical use in clinical and research workflows.
\section{Conclusion}

This work presented a patch-based boundary heatmap framework for corneal layer detection in AS-OCT images, with a particular focus on zero-shot cross-device generalisation. By using native-resolution overlapping patches, the method preserves local anatomical detail, while boundary heatmap regression directly targets the ordered corneal interfaces. Across multiple OCT systems, the proposed method achieved accurate boundary localisation and demonstrated strong zero-shot cross-device generalisation, outperforming representative baseline architectures on unseen devices without device-specific fine-tuning. These results support boundary-focused, resolution-preserving segmentation strategies for robust corneal OCT analysis in clinical settings, particularly across heterogeneous imaging devices and patient populations.

Future work will focus on broader validation and adaptation to the full field of view of different OCT systems, including uncertainty-based masking of poorly visible inner boundaries and post-processing strategies to flag and mitigate failure cases.

\bibliographystyle{unsrt}
\bibliography{references_use}

\end{document}